# Semantic and Contextual Modeling for Malicious Comment Detection with BERT-BiLSTM

Zhou Fang*
Georgia Institute of Technology
Atlanta, USA

Hanlu Zhang
Stevens Institute of Technology
Hoboken, USA

Jacky He
Cornell University
New York, USA

Zhen Qi
Northeastern University
Boston, USA

Hongye Zheng
The Chinese University of Hong
Hong Kong, China

*Abstract*-This study aims to develop an efficient and accurate model for detecting malicious comments, addressing the increasingly severe issue of false and harmful content on social media platforms. We propose a deep learning model that combines BERT and BiLSTM. The BERT model, through pre-training, captures deep semantic features of text, while the BiLSTM network excels at processing sequential data and can further model the contextual dependencies of text. Experimental results on the Jigsaw Unintended Bias in Toxicity Classification dataset demonstrate that the BERT+BiLSTM model achieves superior performance in malicious comment detection tasks, with a precision of 0.94, recall of 0.93, and accuracy of 0.94. This surpasses other models, including standalone BERT, TextCNN, TextRNN, and traditional machine learning algorithms using TF-IDF features. These results confirm the superiority of the BERT+BiLSTM model in handling imbalanced data and capturing deep semantic features of malicious comments, providing an effective technical means for social media content moderation and online environment purification.

## I. INTRODUCTION

With the rapid expansion of the Internet and social media, numerous social platforms have emerged at an unprecedented pace, reflecting the dynamic and exponential growth of digital communication technologies. These platforms leverage complex and dynamic social networks for information dissemination, characterized by fast propagation speed, wide coverage, and significant influence. They greatly facilitate users in obtaining and sharing information. However, this efficient mode of information dissemination also leads to the proliferation of malicious comments and misinformation, posing a global challenge and a serious threat to public opinion and social stability. Malicious comments and misinformation not only spread rapidly but can also trigger negative public emotions, such as fear, disgust, and surprise. Consequently, the identification of false information and malicious content has emerged as a pressing concern that spans various societal domains.

Nevertheless, the detection of malicious comments still faces numerous technical challenges. Issues such as imbalanced data distribution [1], high timeliness requirements [2], and the diversity of language expression significantly limit the generalization and practicality of existing models. Moreover, malicious comments are often highly concealed in the early stages of dissemination. How to utilize deep text semantic features to determine the authenticity of information has become the core challenge in identifying malicious comments. Enhancing the accuracy and efficiency of malicious comment detection is crucial for protecting the public's right to information and maintaining social stability.

Currently, the detection of malicious comments mainly relies on machine learning [3] and deep learning technologies [4], which still have shortcomings in capturing complex semantics and contextual associations. Traditional methods, such as Support Vector Machines (SVM) or TF-IDF, fail to effectively model the deep semantic features of text. Although deep learning models like Convolutional Neural Networks (CNN) [5]and Long Short-Term Memory networks (LSTM) [6] can extract local and temporal information from text, they still have room for improvement when dealing with multimodal data, long texts, and complex contexts.

To tackle these issues, this paper introduces a model for detecting malicious comments, which integrates Bidirectional Encoder Representations from Transformers (BERT) with Bidirectional Long Short-Term Memory networks (BiLSTM). The BERT model, with its powerful pre-trained semantic representation capabilities, can capture the deep semantic features of text [7]. In contrast, the BiLSTM network excels in processing sequential data and can further model the contextual dependencies of text. Compared with existing single models, the proposed BERT-BiLSTM model has unique advantages in extracting semantic and temporal features. By combining the strengths of both, the model can better capture the semantic features of malicious comments and handle challenges related to small sample sizes and imbalanced data. [8].

## II. BACKGROUND

In the early stages, the detection of malicious comments primarily relied on manual verification, which had many drawbacks, such as long processing times, high costs, and low efficiency. These limitations made it difficult to meet the current demand for large-scale malicious comment detection. In recent years, as artificial intelligence and big data technologies have advanced rapidly, automated detection of malicious comments has increasingly become a focal point of research. In conventional machine learning approaches, feature engineering serves as the fundamental methodological basis and has found extensive application in the realm of malicious comment detection. For example, features were selected from three dimensions—content, user, and propagation—to represent the data, and a Support Vector Machine (SVM) was used as the classifier [9]. Additionally, various algorithms, including Logistic Regression, Support Vector Machine, Naïve Bayes, and Decision Trees, were compared, leading to the proposal of a rumor detection method based on user behavior [10]. The specific process is illustrated in Figure 1.

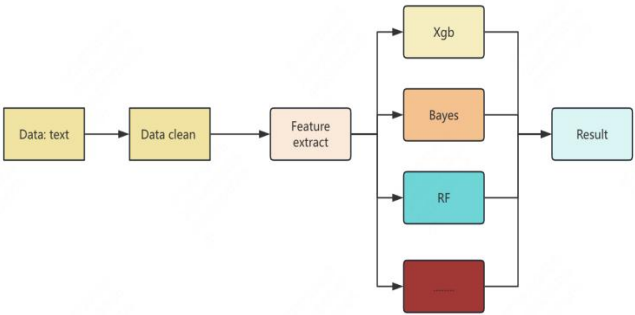

Figure 1. Traditional machine learning Structure

In recent years, deep learning has rapidly evolved into a core component of machine learning. Due to its ability to automatically extract complex features and effectively model data, deep learning models have been widely applied in both research and practical contexts for detecting malicious comments. Notably, three Transformer-based models—BERT, ALBERT, and XLNet—have been introduced for this purpose [11-13].

Building on this foundation, some researchers have begun to explore new methods for malicious comment detection by analyzing the content of news texts. For example, some researchers designed a deep learning model based on bidirectional recurrent neural networks (BiRNN) [14], introduced the Attention mechanism to handle temporal issues in long sequence encoding and decoding, and mapped text data to a low-dimensional real vector space through word embedding [15]. This method performed well in handling the temporal dependencies of long texts but still had shortcomings in capturing the collaborative relationships of complex data.

The Paper extracted deep structural features of public opinion information using Transformer and Multi-head attention mechanisms, integrating document structure and contextual semantic knowledge to improve the accuracy of early identification of false public opinion information [16]. However, the complexity of this model may lead to overfitting risks in different contexts or linguistic features. Researchers proposed a hybrid architecture connecting BERT and RNN, which achieved certain effects in semantic and temporal feature extraction, but RNN may face gradient vanishing problems when processing long texts, limiting its performance in complex language scenarios [17].

In comparison, the dual-model architecture proposed in this paper combines the contextual capture ability of BERT with the temporal modeling capability of BiLSTM, resulting in superior performance in long-text processing and malicious comment detection tasks. The specific process diagram is shown in Figure 2.

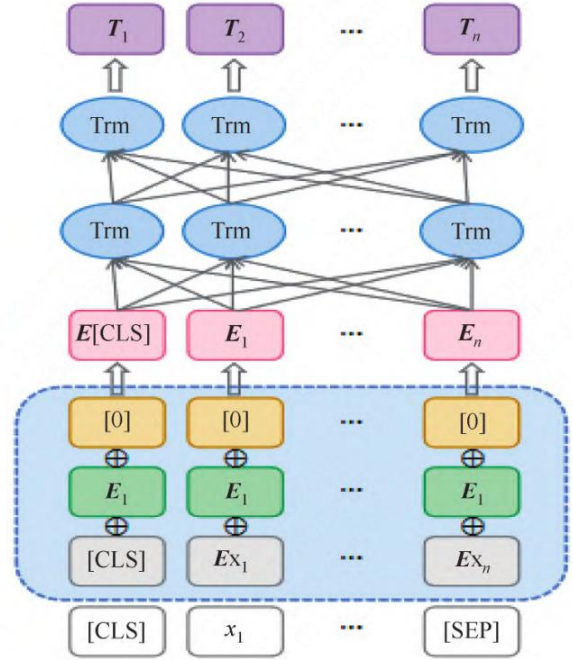

Figure 2. Bert Structure

## III. METHOD

The model begins by encoding the input sentence using BERT, extracting feature vectors from the final layer, and weighting them to create the input for the BiLSTM. Subsequently, the BiLSTM computes the hidden layer vectors in both forward and backward directions, merging them to produce sentence-level feature vectors. Finally, a softmax function is applied to convert these vectors into probabilities [18], facilitating binary classification. This model design leverages BERT's semantic representation capabilities and BiLSTM's capacity to capture contextual information within sentences, thereby enhancing the precision of malicious comment detection. BERT, through its pre-training process on a vast amount of linguistic data, learns rich semantic information and captures bidirectional contextual dependencies within sentences, making it suitable for processing long texts [19]. This combination allows the model to effectively distinguish malicious comments from normal ones, enhancing detection effectiveness. The specific process diagram is shown in Figure 3.

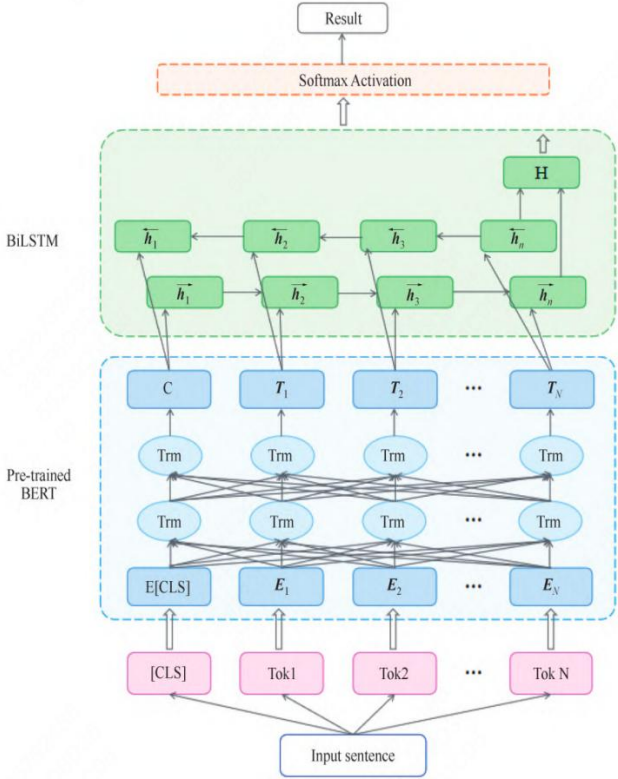

Figure 3. Overall Structure

The model extracts feature vectors from the final layer of BERT's output, incorporating the [CLS] vector. This vector is then subjected to weighting and serves as the input for the BiLSTM model. The detailed calculation is represented by the following formula:

$$a_i = \text{sigmoid}(W_a C_i + b_a), \quad 1 \leq i \leq n$$

Where $C \in R^n$, $W_a \in R^{d_a * n}$; sigmoid() is the activation function; $a_i \in R^{d_a}$; $b_a$ is the bias vector.

The model passes the input vector to the hidden layer. The forward hidden layer vector is $h_{forward}$, and the backward hidden layer vector is $h_{backward}$. The BiLSTM model computes the hidden layer vectors in two opposing directions and subsequently combines the results from these directions. The output vector of the hidden layer at time $i$ is represented as follows:

$$h_i = h_{\text{forward}_i} + h_{\text{backward}_i}$$

Where $h_{forward} \in R^{d_h}$, $h_{backward} \in R^{d_h}$.

Additionally, the model employs the tanh function as the activation function to compute the output of the hidden layer. Here, $d \in \{0,1\}$ denotes the two opposite directions within the hidden layer. The calculation can be expressed as:

$$h_i^d = \tanh(W_h^d a_i + U h_{i-1}^d + b_h^d)$$

All hidden layers $h_i^d$ are concatenated into a vector H, which serves as the final sentence-level feature vector. This feature vector H is subsequently fed into the fully connected layer, where the ReLU function is used as the activation function. The transformation can be expressed as:

$$H_{\text{relu}} = \text{ReLU}(W_{fc} H + b_{fc})$$

Finally, the output layer classifies through the softmax function to calculate the probability of detecting fake news:

$$p(y \mid H, W_s, b_s) = \text{softmax}(W_s H_{\text{relu}} + b_s)$$

### A. Bert Layer

Word2Vec maps text sequences into a dense vector space of word embeddings, after which classification tasks are performed. In contrast, the BERT model employs a pre-training method based on the Transformer architecture [20], which, through training on a large-scale unsupervised corpus, enables the model to effectively adapt to a variety of different downstream tasks [21]. Compared to static word vectors, the BERT model uses a masking mechanism to process text data, which can more accurately interpret the phenomenon of polysemy, thereby enhancing semantic understanding and text classification capabilities. The structure of the BERT model, as shown in Figure 2, relies on a bidirectional Transformer encoder [22], which is implemented through a multi-head attention mechanism and can effectively utilize contextual information to enhance the semantic representation of text.

BERT innovatively introduced two unsupervised tasks: the Masked Language Model (MLM) and Next Sentence Prediction (NSP). The MLM task resembles the cloze test in English, where it predicts masked words using contextual information. The NSP task, on the other hand, determines whether two sentences are consecutive in a classification task. BERT begins with unsupervised training on large-scale corpora to pre-learn character-level, word-level, and sentence-level features effectively. It is then fine-tuned on downstream tasks.

Regarding the model's input and output, the input $E_i$ is a vector composed by adding the token embeddings, segment embeddings, and position embeddings. Each layer of the Transformer encoder outputs the corresponding feature, with each output feature containing attention information for each input. The special token [SEP] is used as a sentence separator, while [CLS] is a special marker that encapsulates the information of the entire sentence, often referred to as the sentence vector, and is commonly used for sentence-level classification.

### B. Bi-LSTM Layer

Long short-term memory is mainly used to solve the problem of gradient disappearance and gradient explosion occurring in long sequence training [23]. In brief, LSTM outperforms ordinary recurrent neural networks when processing longer sequences. The structural diagram of the model is shown in Figure 4. In the input layer, the label information of a sentence is represented by the forward LSTM from the beginning to the end of the sentence and the backward LSTM from the end to the beginning of the sentence. Finally, a comprehensive representation of words is achieved by integrating forward and backward context information. In the input layer, the word vector corresponding to each word in the sentence can be input to the Bi-LSTM layer, the backward hidden layer sequence of the word vectors.

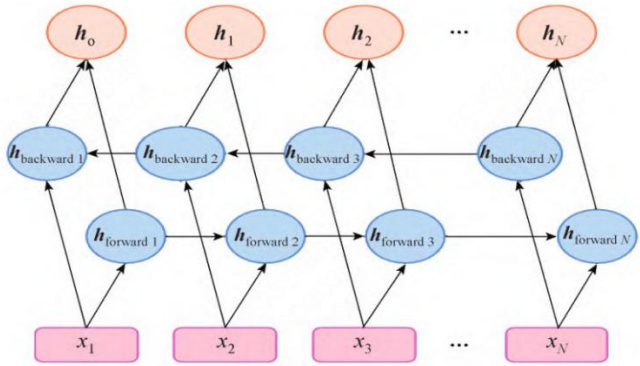

Figure 4. Bi-Lstm Layer

## IV. EXPERIMENT

### A. Dataset

The "Jigsaw Unintended Bias in Toxicity Classification" Dataset [24] serves as an invaluable resource for research in the field of NLP. This dataset was created as part of the Jigsaw initiative in collaboration with Google and was made publicly available on Kaggle. It stems from a series of challenges aimed at reducing toxicity and promoting healthier conversations online.

Under the guidance of ethical standards and research protocols, the dataset was compiled from a diverse range of online platforms. It consists of a large number of comments that have been labeled for toxicity by human raters, ensuring a broad representation of different types of harmful language.

This dataset is especially instrumental for the development and assessment of machine learning models designed to perform toxicity classification on text extracted from online interactions. Such models are pivotal for a variety of applications, including content moderation, user protection, and the enhancement of online community guidelines.

The data provides a robust foundation for researchers to build upon as they work toward creating algorithms capable of accurately identifying and mitigating toxic content. This, in turn, supports the creation of safer online environments and contributes to the broader goal of fostering more positive and inclusive digital communities.

### B. Experiment Results Analysis

In this study, we applied a variety of models to the Jigsaw Unintended Bias in Toxicity Classification dataset, including TF-IDF with XGBoost (TF-IDF+XGB|RF) [25], TextCNN, BERT, and a combination of BERT and LSTM (BERT+LSTM|CNN) [26]. Table 1 is a detailed analysis of the performance of each model.

Table 1. Metric of experiments

| Method | Precision | Recall | Acc |
| --- | --- | --- | --- |
| Tfidf+RF | 0.85 | 0.86 | 0.87 |
| Tfidf+Xgb | 0.88 | 0.89 | 0.89 |
| Textcnn | 0.90 | 0.91 | 0.91 |
| Textrnn | 0.91 | 0.91 | 0.91 |
| Bert | 0.93 | 0.91 | 0.92 |
| Bert+lstm | 0.92 | 0.92 | 0.92 |
| Bert+Cnn | 0.90 | 0.90 | 0.90 |
| Bert+Bilstm | 0.94 | 0.93 | 0.94 |

The experimental results presented in Table 1 provide a comprehensive comparison of various models applied to the dataset. The metrics evaluated include Precision, Recall, and Accuracy, which are crucial for assessing the performance of classification models in detecting toxic comments.

Overview of Model Performance: The TF-IDF+XGB model showed slight improvements with a Precision of 0.88, Recall of 0.89, and Accuracy of 0.89. These results indicate that while these traditional machine learning models are effective, they are outperformed by more advanced deep learning models. The TextCNN model demonstrated better performance with a Precision of 0.90, Recall of 0.91, and Accuracy of 0.91. These results highlight the effectiveness of convolutional and recurrent neural networks in capturing local and sequential features in text data, respectively.

The standalone BERT model showed significant improvements over the baseline models, with a Precision of 0.93, Recall of 0.91, and Accuracy of 0.92. Meanwhile, the BERT+CNN model delivered similar results, with all three metrics at 0.90, suggesting that combining BERT with additional layers, such as CNN or LSTM, can further enhance the model's ability to capture complex linguistic patterns. Among all models, the BERT+BiLSTM model emerged as the best performer, achieving the highest scores: a Precision of 0.94, a Recall of 0.93, and an Accuracy of 0.94. This superior performance likely stems from BERT's deep contextual understanding combined with BiLSTM's ability to capture dependencies in both forward and backward directions.

## V. CONCLUSION

The results clearly demonstrate the advantages of deep learning models, particularly those that leverage pre-trained language representations like BERT. The combination of BERT with BiLSTM proved to be the most effective approach, achieving the highest metrics across Precision, Recall, and Accuracy. This suggests that integrating BERT's contextual awareness with the temporal dynamics captured by BiLSTM can significantly improve the detection of toxic comments in online platforms. Future work could explore further optimizations and the application of these models to real-world scenarios to enhance online safety and content moderation.

Future work could focus on optimizing the BERT+BiLSTM model, such as adjusting hyperparameters, refining the pre-training process, or incorporating additional contextual features. Beyond malicious comments, the model could also be adapted to detect other forms of harmful content, such as hate speech, cyberbullying, and misinformation. Expanding the model's applicability to a broader range of online safety issues would enhance its practical value in real-world content moderation systems. Moreover, the integration of the BERT+BiLSTM model into existing content moderation systems could provide a scalable solution for platforms seeking to automatically identify and mitigate harmful content, thereby fostering healthier online communities.

Furthermore, while the BERT+BiLSTM model shows promise, it is crucial to consider the ethical implications of automated content moderation. The model's decisions should be transparent and explainable to users, and there should be mechanisms in place for human oversight to address any potential biases or errors in the model's predictions. Ensuring that the model operates within ethical guidelines and respects user rights is crucial as we move towards more automated forms of content moderation.